\documentclass[10pt]{article}
\usepackage[utf8]{inputenc}
\usepackage[T1]{fontenc}
\usepackage{objdream_preprint}
\usepackage[round,authoryear]{natbib}
\usepackage{graphicx}
\usepackage{booktabs}
\usepackage{tabularx}
\usepackage{multirow}
\usepackage{amsmath,amssymb}
\usepackage{microtype}
\usepackage{enumitem}
\usepackage{float}
\usepackage{placeins}
\usepackage{url}
\usepackage{xcolor}
\newcommand{\ours}{ObjDream}
\newcommand{\libero}{LIBERO}

\usepackage{hyperref}
\definecolor{paperblue}{HTML}{357CBC}
\hypersetup{
    colorlinks=true,
    citecolor=paperblue,
    linkcolor=paperblue,
    urlcolor=paperblue,
    bookmarksnumbered=true,
    pdfsubject={Robotic manipulation; preprint},
    pdftitle={ObjDream: Guiding Vision-Language-Action Models with Object-Level Understanding and Future State Prediction}
}
\title{ObjDream: Guiding Vision-Language-Action Models
with Object-Level Understanding and Future State Prediction}
\date{}

\newcommand{\AuthorOneName}{Huailiang Ma}
\newcommand{\AuthorOneAffiliations}{1,2}

\newcommand{\AuthorTwoName}{Yuxuan Li}
\newcommand{\AuthorTwoAffiliations}{3}

\newcommand{\AuthorThreeName}{Ziyi Zhao}
\newcommand{\AuthorThreeAffiliations}{1,2}

\newcommand{\AuthorFourName}{Xinkai Wang}
\newcommand{\AuthorFourAffiliations}{1,2}

\newcommand{\AuthorFiveName}{Mingyu Li}
\newcommand{\AuthorFiveAffiliations}{1}

\newcommand{\AuthorSixName}{Xinyu Guo}
\newcommand{\AuthorSixAffiliations}{4}

\newcommand{\AuthorSevenName}{Lifeng Zhu}
\newcommand{\AuthorSevenAffiliations}{1}

\newcommand{\AuthorEightName}{Aiguo Song}
\newcommand{\AuthorEightAffiliations}{1}

\newcommand{\AuthorNineName}{Cewu Lu}
\newcommand{\AuthorNineAffiliations}{2,3}

\newcommand{\AuthorTenName}{Jianhua Sun}
\newcommand{\AuthorTenAffiliations}{2,3}

\newcommand{\AffiliationOne}{Southeast University}

\newcommand{\AffiliationTwo}{Shanghai Innovation Institute}

\newcommand{\AffiliationThree}{Shanghai Jiao Tong University}

\newcommand{\AffiliationFour}{Institute of Automation, Chinese Academy of Sciences}

\newcommand{\ContactLine}{\\[4pt]{\normalfont\small\textsuperscript{*}Corresponding author.}}

\author{\AuthorOneName\textsuperscript{\AuthorOneAffiliations}\quad
  \AuthorTwoName\textsuperscript{\AuthorTwoAffiliations}\quad
  \AuthorThreeName\textsuperscript{\AuthorThreeAffiliations}\quad
  \AuthorFourName\textsuperscript{\AuthorFourAffiliations}\quad
  \AuthorFiveName\textsuperscript{\AuthorFiveAffiliations}\\[4pt]
  \AuthorSixName\textsuperscript{\AuthorSixAffiliations}\quad
  \AuthorSevenName\textsuperscript{\AuthorSevenAffiliations}\quad
  \AuthorEightName\textsuperscript{\AuthorEightAffiliations}\quad
  \AuthorNineName\textsuperscript{\AuthorNineAffiliations}\quad
  \AuthorTenName\textsuperscript{\AuthorTenAffiliations,*}\\[7pt]
  {\normalfont\small
  \textsuperscript{1}\AffiliationOne
  \quad
  \textsuperscript{2}\AffiliationTwo
  \quad
  \textsuperscript{3}\AffiliationThree
  }\\[3pt]
  {\normalfont\small
  \textsuperscript{4}\AffiliationFour
  }\ContactLine
}

\hypersetup{
  pdfauthor={
    \AuthorOneName, \AuthorTwoName, \AuthorThreeName,
    \AuthorFourName, \AuthorFiveName, \AuthorSixName,
    \AuthorSevenName, \AuthorEightName, \AuthorNineName,
    \AuthorTenName
  }
}

\newcommand{\projecturl}{https://huailiangma.github.io/objdream/}
\newcommand{\institutionmasthead}{\noindent\makebox[\textwidth][l]{\includegraphics[width=0.22\textwidth,height=28pt,keepaspectratio]{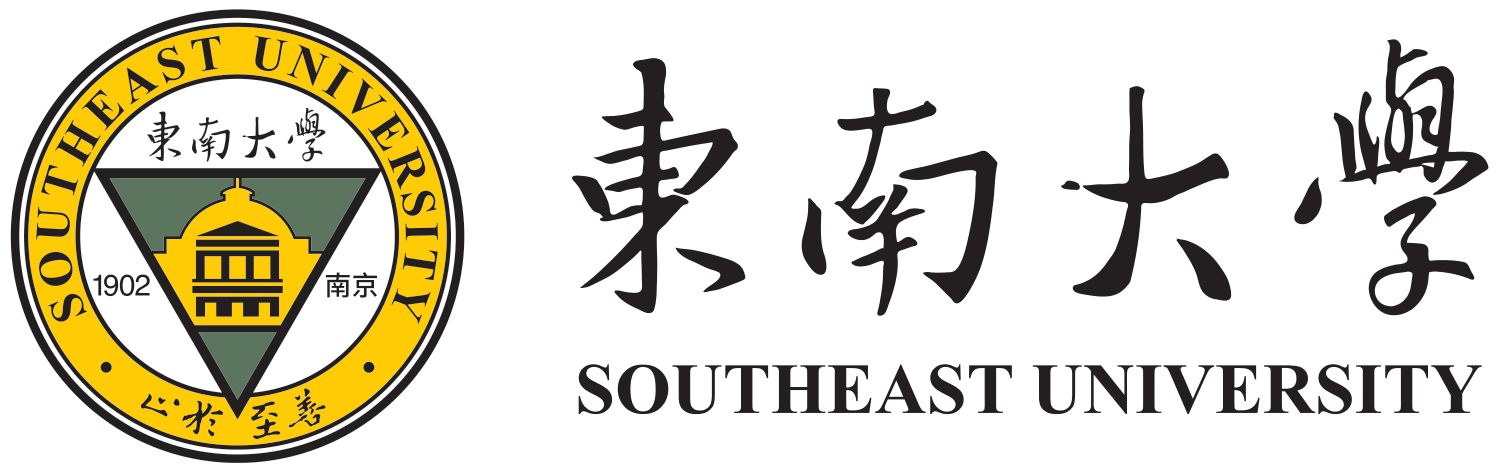}\hspace{10pt}\includegraphics[width=0.22\textwidth,height=28pt,keepaspectratio]{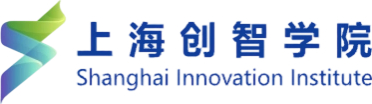}\hspace{10pt}\includegraphics[width=0.22\textwidth,height=28pt,keepaspectratio]{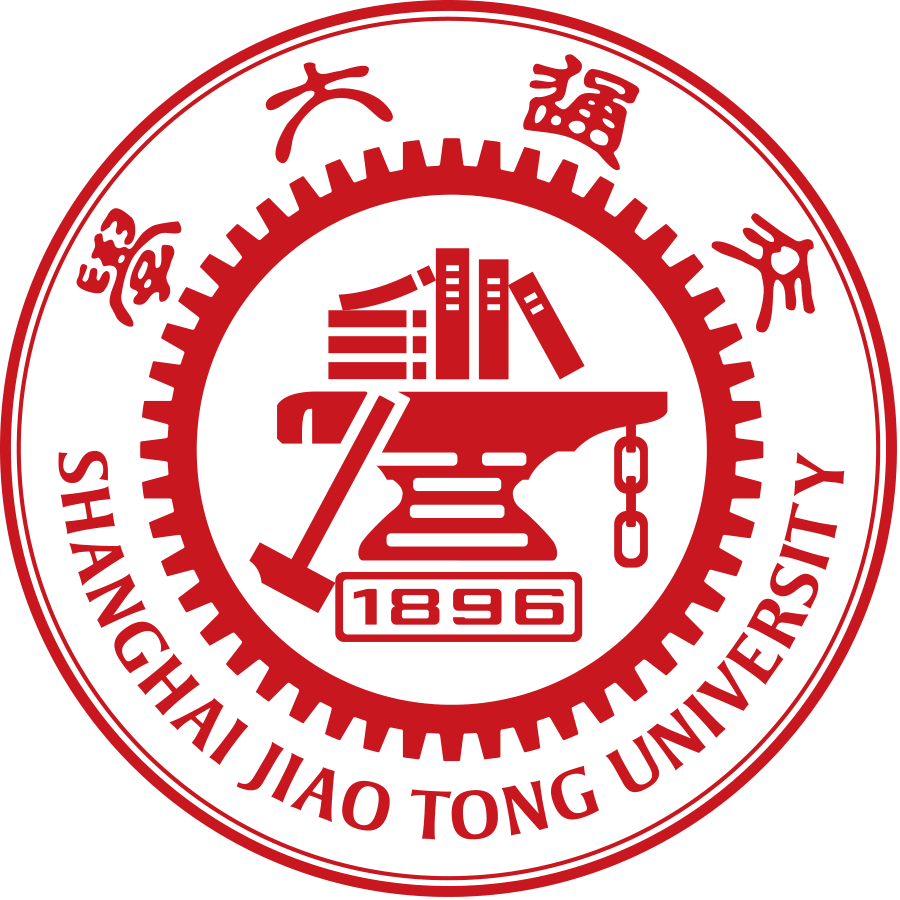}\hspace{10pt}\includegraphics[width=0.22\textwidth,height=28pt,keepaspectratio]{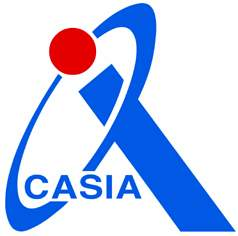}}\par
  \vspace{4pt}
  {\color{black!55}\hrule height 0.5pt}}

\makeatletter
\renewcommand{\@maketitle}{\vbox{\hsize\textwidth
    \linewidth\hsize
    \setlength{\parskip}{0pt}\institutionmasthead
    \vskip 11pt
    \centering
    \makebox[\textwidth][c]{\begin{minipage}{1.12\textwidth}
        \centering
        \setlength{\parskip}{0pt}
        {\fontsize{16}{18}\selectfont\bfseries\@title\par}\vspace{9pt}
        {\normalfont\normalsize
          \begin{tabular}{@{}c@{}}
            \@author
          \end{tabular}\par}\end{minipage}}\par
    \vskip 7pt
  }}
\makeatother

\newcommand{\bodygeometry}{\newgeometry{textwidth=5.5in,textheight=9in,top=81.86bp,
    headheight=12pt,headsep=25pt,footskip=30pt}}

\AtBeginDocument{\newgeometry{textwidth=5.5in,textheight=9.3in,top=50bp,
    headheight=12pt,headsep=25pt,footskip=30pt}}

\begin{document}
\maketitle

\begin{abstract}
Vision-language-action (VLA) models have emerged as a powerful paradigm for robotic manipulation, enabling robots to generate actions from visual observations and language instructions.
However, most existing VLA models still operate on image-level visual representations inherited from pretrained vision-language backbones, while lacking fine-grained object understanding for manipulation.
As a result, their robustness can degrade under out-of-distribution object layouts, clutter, and background variations.
To address these limitations, we introduce \ours{}, an object-centric VLA framework centered on object-level understanding and future object state prediction.
\ours{} uses a spatio-temporal object encoder to organize visual observations into temporal object slots, where persistent object identity and dynamic object state are represented through separate channels.
These structured object representations are inserted into the token sequence of the vision-language model (VLM), where identity-guided routing modulates VLM attention to the corresponding object tokens.
In addition, an action-conditioned object dynamics module predicts compact future object states relevant to manipulation during training, explicitly supervising object state evolution under robot actions without requiring future prediction at inference.
Experiments on \libero{}, \libero-Plus, and RoboTwin~2.0 show that \ours{} improves generalization under diverse distribution shifts and scene perturbations.
Real-world robot experiments further demonstrate that \ours{} transfers to physical manipulation settings with novel object arrangements and background changes. \href{\projecturl}{Project page}.
\end{abstract}

\begin{figure}[H]
    \centering
    \includegraphics[width=0.99\linewidth]{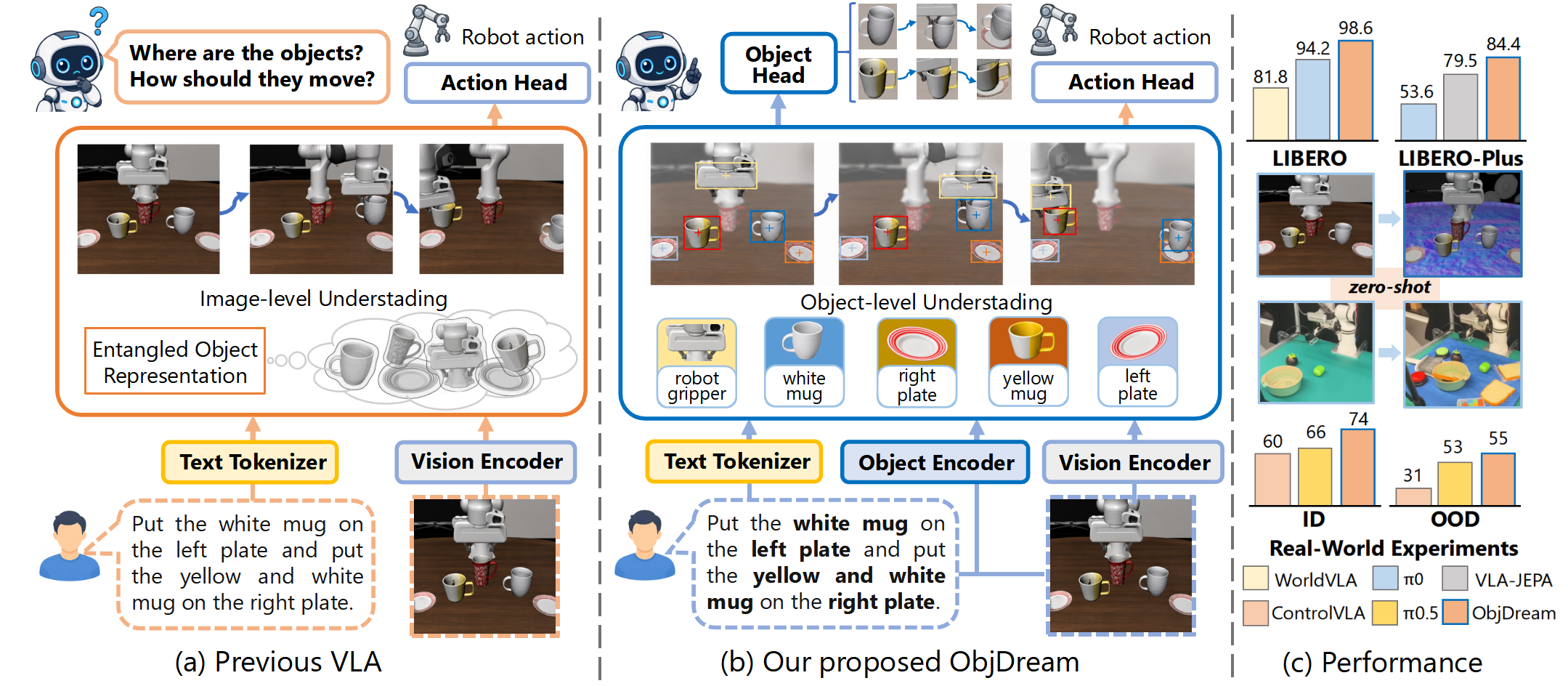}
    \caption{\textbf{Overview of \ours{}.}
    We introduce \ours{}, an object-centric VLA framework for robust manipulation.
    Unlike prior VLA models (a), which primarily rely on image-level understanding and entangled object representations, \ours{} (b) introduces explicit object-level representations and future object state prediction.
    This object-centric formulation improves manipulation robustness, leading to stronger performance than baseline methods (c).}
    \label{fig1_overview}
\end{figure}

\bodygeometry

\section{Introduction}
Robust robot policies must reliably manipulate objects in out-of-distribution environments, where their appearance, position, and surrounding context may differ from those in the training data.
VLA models have made rapid progress in generalist robot control by scaling pretrained vision-language backbones, robot data, and action decoders~\citep{brohan2023rt2,kim2024openvla,black2024pi0,gr00t2025,wang2026qwen}.
To improve generalization, recent work has sought to preserve knowledge acquired during pretraining~\citep{guo2026priorvla,hancock2026actions}, use high-level plans to guide low-level control~\citep{li2025hamster}, or anticipate scene changes through future visual or latent-state prediction~\citep{zhang2025dreamvla,sun2026vla}.
Although these approaches improve generalization to some extent, most VLA models still lack fine-grained object-level understanding. Their visual representations, inherited from VLMs, remain largely image-level, leaving the identity and spatial state of task-relevant objects entangled with appearance and surrounding context. This entanglement can weaken out-of-distribution robustness under shifts in object appearance, spatial layout, and background context.

Object-centric approaches offer a natural way to strengthen object-level understanding by explicitly representing individual objects and their properties.
Prior work incorporates object-aware visual representations to help policies focus on task-relevant regions~\citep{zhu2023viola,li2025controlvla,liang2026pixelvla}.
Other methods extend object representations to relational reasoning~\citep{hanyu2025slotvla}, object-level memory~\citep{chung2026rethinking}, and long-horizon manipulation~\citep{yang2026lilo}.
Although these approaches demonstrate the value of object-level representations, they often encode persistent object identity and time-varying object state in the same slot or token. As appearance and geometry change, this coupling can obscure persistent identity cues by mixing them with dynamic object state.
Recent dual-state slot models show that separating identity from dynamic appearance can improve temporal consistency in video perception by keeping identity information distinct from time-varying visual content~\citep{tran2026dual}.
Building on this insight, we investigate whether persistent identity references can be represented alongside dynamic object states and further used to guide VLM attention for language-conditioned manipulation.

Beyond modeling the current visual observation, predictive models support manipulation by anticipating future observations or object states~\citep{ferraro2025focus,chen2025goalvla,villar2025playslot}.
The choice of prediction target is important: reconstructing complete images requires modeling texture and background details that may be irrelevant to control~\citep{zhang2025dreamvla}, while goal-image methods such as Goal-VLA introduce an additional synthesis stage before estimating the target object pose~\citep{chen2025goalvla}.
In contrast, object-centric world models show that scene evolution can be modeled using compact object representations~\citep{jeong2025object}.
Motivated by this trade-off, we use action-conditioned prediction of compact future object states, including geometry and visibility, as auxiliary supervision for VLA training without requiring future object state prediction at inference.

To this end, as illustrated in Fig.~\ref{fig1_overview}, we introduce \ours{}, an object-centric vision-language-action framework for improving manipulation generalization. The architecture of \ours{} builds on a pretrained VLM backbone and integrates a spatio-temporal object encoder, an object-level forward dynamics module, and a flow-matching action expert. Specifically, the spatio-temporal object encoder transforms visual observations into structured object slots, where persistent object identity and dynamic object state are represented through separate channels. These factorized object representations are mapped into the VLM hidden space and inserted into the token sequence, where identity-guided routing modulates attention to the corresponding object tokens. The forward dynamics module predicts future object states during training, providing auxiliary supervision for future object state evolution under robot actions without requiring future prediction at inference. The object encoder is pretrained with automatically generated offline object supervision and frozen during VLA training. We optimize \ours{} with a joint objective combining action imitation with current and future object state supervision. Extensive evaluations on LIBERO, LIBERO-Plus, RoboTwin~2.0, and real-world robot experiments demonstrate competitive manipulation performance across diverse scene perturbations.

\noindent The main contributions of this paper are listed below:
\begin{itemize}[leftmargin=2em, topsep=0.25em, itemsep=0.25em, parsep=0pt]
    \item We present \ours{}, an object-centric VLA framework centered on object-level understanding and future object state prediction to improve generalization under perturbations. \ours{} introduces a spatio-temporal object encoder that explicitly represents persistent object identity and dynamic object state through separate channels. The resulting representations are integrated into the VLM hidden space with a novel identity-guided routing mechanism for robust language-conditioned manipulation.

    \item We introduce an action-conditioned object-level forward dynamics module for modeling object evolution during manipulation. It predicts compact future object states during training to provide direct auxiliary object-level supervision without requiring future object state prediction at inference. We further model inter-object relations from object features and pairwise geometry to provide structured context for action generation.

    \item Extensive evaluations on LIBERO, LIBERO-Plus, RoboTwin~2.0, and real-world robot experiments demonstrate competitive manipulation performance across diverse scene and environmental perturbations. Ablations further assess the effects of the key architectural components.

\end{itemize}

\section{Related Work}
\paragraph{Vision-language-action models.}
VLA models combine pretrained language or vision-language backbones with action decoders to learn general-purpose manipulation from large-scale robot demonstrations~\citep{brohan2023rt1,brohan2023rt2,openx2023embodiment,octo2024,kim2024openvla,black2024pi0,pi05}. Recent work improves VLA performance through stronger action decoders and fine-tuning strategies~\citep{chi2023diffusionpolicy,zhao2023learning,kim2025fine,hou2025dita,gr00t2025,qu2025spatialvla}, larger-scale cross-embodiment training~\citep{cai2026xiaomi,team2026xiaomi,wu2026pragmatic,wu2026foundation,wang2026qwen}, and better preservation or exploitation of pretrained visual-semantic and spatial priors~\citep{guo2026priorvla,li2026spatial,wang2026lamp}. Predictive objectives further use future images, video tokens, subgoal images, or latent states to provide temporal supervision for manipulation~\citep{zhao2025cot,zhang2025dreamvla,cen2025worldvla,cen2025rynnvla,li2026fomovla}. Despite these advances, existing VLA models still largely operate on dense image-level visual representations, where object-level structure is not explicitly organized for manipulation generalization. \ours{} addresses this limitation by enhancing VLA policies with object-level understanding and future object state supervision.

\paragraph{World models for robot control.}
World models learn action-conditioned environment dynamics and support robot control by planning or learning policies from predicted futures in observation or latent spaces~\citep{hafner2019planet,wu2023daydreamer}. Recent work extends this paradigm through language-conditioned video generation~\citep{du2023learning,zhou2024robodreamer}, pretrained visual latent dynamics~\citep{zhou2024dino,assran2025v}, and world-action or unified video-action models that couple future prediction with executable actions~\citep{ye2026world,li2026latent}. These approaches provide dynamics supervision by modeling anticipated scene evolution. However, prediction in pixel or video space requires modeling high-dimensional visual details and can introduce training or inference overhead. Recent approaches therefore explore more efficient formulations, including avoiding explicit future imagination at inference or exposing compact representations of scene changes~\citep{yuan2026fast,bai2026bridge}. \ours{} follows this direction by using compact future object state prediction as supervision during training, while directly generating actions at inference without explicit future object state decoding.

\paragraph{Object-centric representations for manipulation.}
Object-centric representation learning decomposes visual scenes into entity-level latent variables and has been widely studied in scene decomposition, temporal tracking, and structured dynamics~\citep{locatello2020object,kipf2021conditional,wu2022slotformer}.
In robotic manipulation, object-level representations have been used to improve sample efficiency and enhance visuomotor attention and robust policy conditioning through object proposals, grounded geometry, object-centric prompting, and physically grounded object concepts~\citep{zhu2023viola,vo2025clutter,li2025object,wei2026physically}, enable efficient VLA learning through compact object-centric tokenization~\citep{bendikas2025focusing}, and support structured control through object-centric dynamics and relational representations~\citep{ferraro2025focus,mosbach2024sold,hanyu2025slotvla,yang2026lilo}.
Complementary approaches use structured affordance representations to provide action-relevant object grounding, interaction localization, and geometric guidance for VLA policies~\citep{yu2026affordancevla}.
Recent predictive approaches further incorporate object-aware future representations, masks, or addressable slot states into world-action learning~\citep{wang2026oflow,yu2026maskwam,liu2026oa}.
Recent dual-state slot models separate per-frame appearance from persistent object identity, highlighting the value of disentangling identity from changing visual content in object-centric representations~\citep{tran2026dual}.
From this perspective, \ours{} combines persistent identity references with dynamic object states and uses identity-guided routing to provide object-specific context for language-conditioned manipulation.

\begin{figure}
    \centering
    \includegraphics[width=1\linewidth]{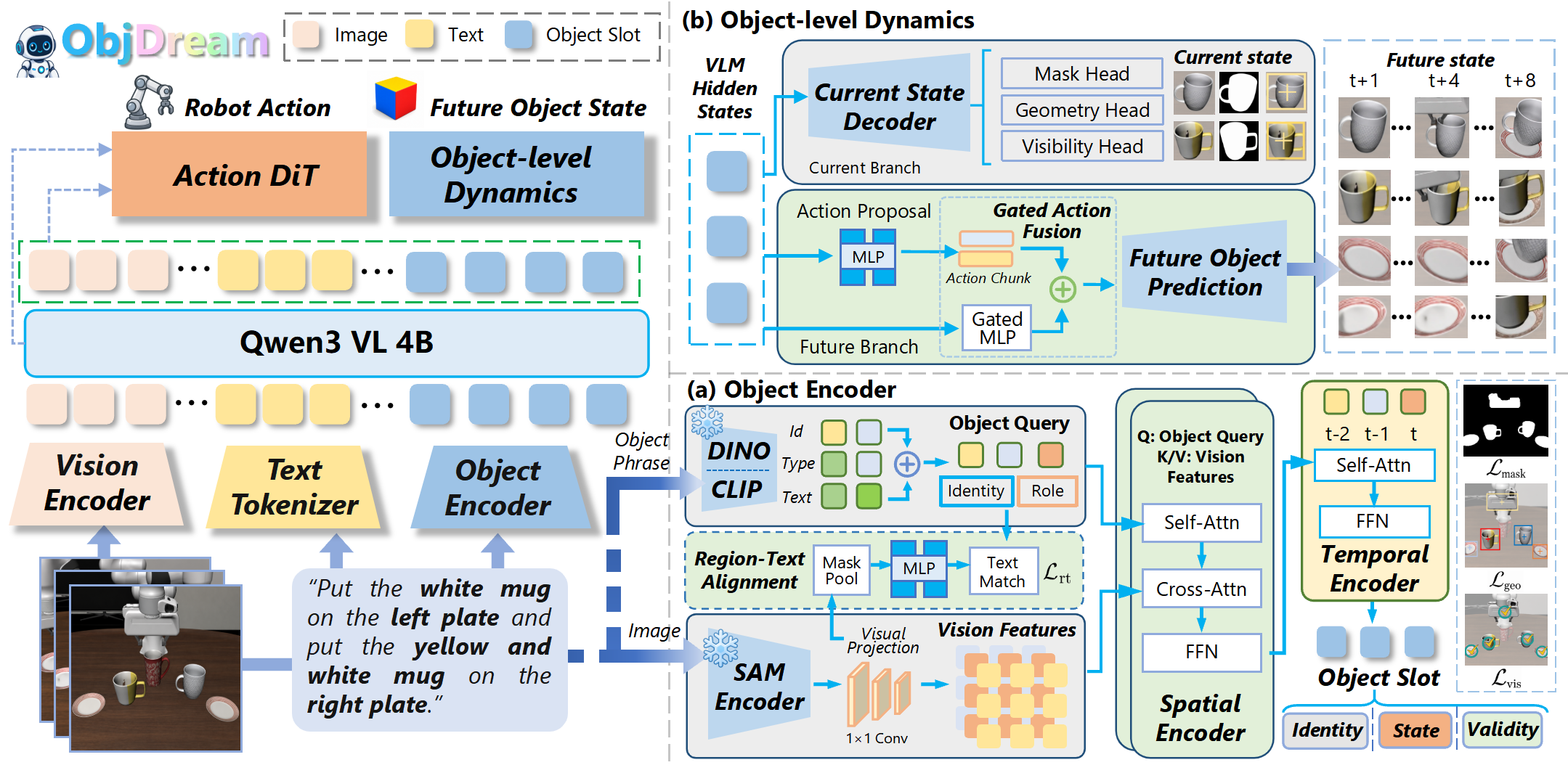}
    \caption{\textbf{Overview of the \ours{} architecture.}
    The model integrates three main components: (1) a spatio-temporal object encoder that constructs factorized object slots by separately representing persistent object identity and time-varying object state; (2) an object-level dynamics module that decodes current object states and predicts action-conditioned future object states; and (3) a flow-matching action expert that predicts robot action chunks. \ours{} enhances object-level understanding through factorized object representations and future object state modeling, enabling more generalizable language-conditioned manipulation under object and scene variations.}
    \label{fig2_architecture}
\end{figure}

\section{Method}
\label{sec:method}
Following the VLA policy formulation in Sec.~\ref{sec:VLA Policy Formulation}, Fig.~\ref{fig2_architecture} illustrates the architecture of \ours{}.
\ours{} combines a spatio-temporal object encoder, an object dynamics module, and a flow-matching action expert.
The encoder produces structured object slots that are mapped to VLM object tokens, while identity-guided routing uses cached identity references to modulate attention to these tokens (Sec.~\ref{Object-Centric Tokenization and VLM Routing}).
The object dynamics module provides future object state supervision during training (Sec.~\ref{sec:Object-Level Forward Dynamics Module}), and the action expert generates action chunks.

\subsection{VLA Policy Formulation}
\label{sec:VLA Policy Formulation}
We formulate \ours{} as an object-centric VLA policy that predicts continuous action chunks from language instructions, visual observations, proprioceptive states, and object-level embeddings. At control step $t$, the policy receives a language instruction $x$, a current multi-view observation $I_t$, and proprioceptive state $s_t$. The VLM consumes all policy views in $I_t$, while the object encoder operates on a designated external or third-person view with causal history $\mathbf{I}_t=(I_{t-T+1}^{\mathrm{enc}},\ldots,I_t^{\mathrm{enc}})$.
At episode initialization, a phrase parser extracts a task-relevant phrase set $P_0$ from the instruction $x$ and initial observation $I_0$.
A spatio-temporal object encoder $E_\omega$ then produces structured object slots $O_t=E_\omega(\mathbf{I}_t,P_0)$.
The VLA backbone $f_\psi$ encodes these inputs into a multimodal context $C_t$, and the action expert parameterizes the conditional distribution over an action chunk $A_t=[a_t,\ldots,a_{t+H_a-1}]$:
\begin{equation}
C_t=f_\psi(x,I_t,s_t,O_t),\qquad p_\theta(A_t\mid x,I_t,s_t,O_t)=p_\theta(A_t\mid C_t).
\label{eq:vla_policy}
\end{equation}
Here, $H_a$ denotes the action horizon. In our implementation, $p_\theta$ is instantiated by a flow-matching action expert~\citep{lipman2023flow,black2024pi0}. Given a demonstrated action chunk $A_t^\star$, Gaussian noise $Z_t\sim\mathcal{N}(0,I)$, and flow time $\tau\sim\mathcal{U}[0,1]$, the action loss is
\begin{equation}
A_t^\tau=(1-\tau)Z_t+\tau A_t^\star,\qquad
\mathcal{L}_{\mathrm{act}}
=
\mathbb{E}\!\left[
\left\|v_\theta(A_t^\tau,\tau;C_t)-(A_t^\star-Z_t)\right\|_2^2
\right].
\label{eq:action_loss}
\end{equation}
At inference, the learned velocity field is integrated to generate an action chunk, which is executed in a receding-horizon manner.

\subsection{Object-Centric Tokenization and VLM Routing}
\label{Object-Centric Tokenization and VLM Routing}

\paragraph{Spatio-temporal object encoder.}
\ours{} uses a spatio-temporal object encoder $E_\omega$ to construct factorized object slots from the causal image window $\mathbf{I}_t$ and the cached phrase set $P_0$.
At episode initialization, Qwen3-VL-8B~\citep{bai2025qwen3} processes the instruction $x$ and initial observation $I_0$ once to extract task-relevant object phrases, which are cached for all subsequent control steps.
Query ordering is fixed across frames, with reserved robot queries and padding to $N_s$ slots.
For each phrase $p_i$, a frozen CLIP text encoder~\citep{clip} produces $f_i$, and a learnable projection gives $\iota_i=\operatorname{Normalize}(P_{\mathrm{sem}}(f_i))$.
The object query is initialized as
\begin{equation}
q_i^{(0)}=P_{\mathrm{id}}(\iota_i)+P_{\mathrm{text}}(f_i)+E_{\mathrm{type}}(\chi_i),
\end{equation}
where $P_{\mathrm{id}},P_{\mathrm{text}}$ are learnable projections and $E_{\mathrm{type}}$ embeds the query type $\chi_i$.
A frozen DINOv2 encoder~\citep{dinov2} extracts a visual reference $d_i$ from the initial object region predicted by the encoder, which is cached throughout the episode.
Robot queries use learned identities and type priors without the direct text term.

A spatial query module applies self-attention and cross-attention to learnably projected features from a frozen SAM2 image backbone~\citep{sam2}, followed by causal temporal encoding.
The encoder outputs
\begin{equation}
O_t=E_\omega(\mathbf{I}_t,P_0)=\left\{u_{i,t}^{\mathrm{obj}}\right\}_{i=1}^{N_s}, \qquad
u_{i,t}^{\mathrm{obj}}=(e_i^{\mathrm{id}},e_{i,t}^{\mathrm{state}},\nu_{i,t}).
\end{equation}
The persistent identity $e_i^{\mathrm{id}}=[\iota_i;d_i;r_i]$ combines semantic and visual references with a role embedding $r_i$, while $e_{i,t}^{\mathrm{state}}=[c_{i,t}^{\mathrm{app}};g_{i,t}^{\mathrm{geo}};\tau_t]$ encodes dynamic appearance, geometry, and temporal information.
Slot validity $\nu_{i,t}$ masks inactive or padded slots and is distinct from predicted visibility.
During pretraining, regional features obtained by mask pooling from the projected SAM2 features, extracted before query cross-attention, are aligned with frozen CLIP text features.
The objective is described in Sec.~\ref{sec:training_objective}.

\paragraph{Factorized object injection.}
We map identity and dynamic state components separately into the VLM hidden space:
\begin{equation}
h_{i,t}^{\mathrm{obj}}=
\left[
P_{\mathrm{id}}^{\mathrm{inj}}(e_i^{\mathrm{id}});
P_{\mathrm{obj}}\left(
[P_{\mathrm{state}}(e_{i,t}^{\mathrm{state}});\phi_\nu(\nu_{i,t})]
\right)
\right].
\end{equation}
Here, $P_{\mathrm{id}}^{\mathrm{inj}}$, $P_{\mathrm{state}}$, and $P_{\mathrm{obj}}$ are lightweight projections, and $\phi_\nu$ embeds slot validity.
The factorization is imposed at the encoder output and initial object-token construction; subsequent VLM contextualization is unrestricted, while the cached identity reference $e_i^{\mathrm{id}}$ remains separate for routing.
Attention keys and values are computed from the full contextualized token.

\paragraph{Relation and prospective object tokens.}
Beyond individual object tokens, \ours{} introduces a relation token and $N_p=2$ prospective object tokens:
\begin{equation}
h_t^{\mathrm{rel}}
=
\Phi_{\mathrm{rel}}
\left(
[\mathrm{Pool}_{i}(h_{i,t}^{\mathrm{obj}});
\mathrm{Pool}_{i,j}(\Delta g_{ij,t})]
\right),
\qquad
h_{k,t}^{\mathrm{pro}}
=
\Phi_{\mathrm{pro}}
\left(
[h_{i_k,t}^{\mathrm{obj}};h_t^{\mathrm{rel}}]
\right).
\end{equation}
Here, $\Phi_{\mathrm{rel}}$ aggregates object features and pairwise geometry into a compact global relational context, while $\Phi_{\mathrm{pro}}$ combines this context with individual object features.
$\mathrm{Pool}$ denotes aggregation over valid slots or valid slot pairs, and $\Delta g_{ij,t}$ encodes the relative geometry between slots $i$ and $j$.
The indices $i_k$ correspond to the top-$N_p$ valid slots ranked by a learned slot importance gate.
These prospective tokens are optimized through the action objective to provide object-specific context for action generation.
Together, the tokens form the object block
\begin{equation}
X_t^{\mathrm{obj}}
=
[h_{1,t}^{\mathrm{obj}},\ldots,h_{N_s,t}^{\mathrm{obj}},
h_t^{\mathrm{rel}},
h_{1,t}^{\mathrm{pro}},\ldots,h_{N_p,t}^{\mathrm{pro}}].
\end{equation}
The block is inserted between the multimodal prefix and the action-generation prompt.
The attention mask retains causal attention and additionally permits bidirectional attention within the object block, excluding invalid and padded slots as keys.

\paragraph{Identity-guided object routing.}
For the base object token associated with slot $i$ at VLM position $j_i$, query position $p$, layer $\ell$, and attention head $m$, let $\bar h_p^\ell$ denote the normalized query hidden state.
Using the cached identity reference $e_i^{\mathrm{id}}$, we compute an identity matching score $b_{pi}^{\ell}$ and a head-specific gate $\eta_p^{\ell,m}$:
\begin{equation}
b_{pi}^{\ell}
=
\cos\!\left(U^\ell\bar h_p^\ell,A^\ell e_i^{\mathrm{id}}\right),
\qquad
\eta_p^{\ell,m}
=
\eta_{\max}
\sigma\!\left((w_g^{\ell,m})^\top\bar h_p^\ell+c_g^{\ell,m}\right).
\label{eq:identity_matching}
\end{equation}
Here, $U^\ell$ and $A^\ell$ are learnable projections shared across heads, and $\cos(\cdot,\cdot)$ denotes cosine similarity between their $\ell_2$-normalized outputs. The parameters $w_g^{\ell,m}$ and $c_g^{\ell,m}$ define the head-specific gate $\eta_p^{\ell,m}\in(0,\eta_{\max})$.
For each valid base slot ($\nu_{i,t}=1$), the backbone attention logit is augmented as
\begin{equation}
s_{p j_i}^{\ell,m}
=
\frac{(q_p^{\ell,m})^\top k_{j_i}^{\ell,m}}{\sqrt{d_k}}
+
\eta_p^{\ell,m}b_{pi}^{\ell}
+
M_{p j_i},
\label{eq:identity_guided_routing}
\end{equation}
where $q_p^{\ell,m}$ and $k_{j_i}^{\ell,m}$ are the standard backbone query and key, $d_k$ is the key dimension, and $M$ is the attention mask.
The cached identity affects only the additive bias; standard keys and values are computed from the full contextualized tokens.
All other logits remain unchanged and are normalized by the same softmax.

\subsection{Object-Level Forward Dynamics Module}
\label{sec:Object-Level Forward Dynamics Module}

\paragraph{Action-conditioned object forward dynamics.}
Let $o_{i,t}$ denote the final VLM hidden state of the object token associated with slot $i$.
An auxiliary head predicts an $H_a$-step action chunk $\tilde A_t$ from mean-pooled, stop-gradient features of valid slots, without using the action expert's output.
We construct a position-aware action embedding $e_t^a$ and action conditioning gates $\beta_{i,t}$:
\begin{equation}
e_t^a=\frac{1}{H_a}\sum_{h=1}^{H_a}P_a([\mathrm{sg}(\tilde A_{t,h});p_h]), \qquad
\beta_{i,t}=\sigma(P_\beta(\mathrm{sg}(o_{i,t}))).
\end{equation}
Here, $p_h$ is a learned temporal embedding, $P_a$ and $P_\beta$ are trainable networks, $\sigma$ is sigmoid, and $\mathrm{sg}$ denotes stop-gradient.
The action conditioning gate $\beta_{i,t}$ is distinct from the slot importance gate.
During warmup, demonstrated actions $A_t^\star$ replace $\tilde A_t$ in the conditioning path.
For valid slots, current states are decoded directly, while future prediction is restricted to non-robot object slots:
\begin{equation}
\hat Y_{i,t}^{\mathrm{cur}}=D_{\mathrm{cur}}(o_{i,t}), \qquad
\hat Y_{i,t+1:t+H_f}^{\mathrm{fut}}=D_{\mathrm{fut}}(o_{i,t}+\gamma\beta_{i,t}e_t^a).
\end{equation}
Here, $D_{\mathrm{cur}}$ decodes current states, $\gamma$ is a fixed residual scale, and $D_{\mathrm{fut}}$ predicts all $H_f$ future horizons in parallel, producing horizon-indexed geometry, visibility, and mask logits; robot slots are excluded since their motion is represented by the action chunk.
The auxiliary branch is skipped at inference, when the policy directly generates action chunks.

\subsection{Training Objectives}
\label{sec:training_objective}

\paragraph{Object encoder pretraining.}
We pretrain $E_\omega$ using structured object-level supervision:
\begin{equation}
\mathcal{L}_{\mathrm{enc}}=\lambda_{\mathrm{mask}}\mathcal{L}_{\mathrm{mask}}+\lambda_{\mathrm{geo}}\mathcal{L}_{\mathrm{geo}}+\lambda_{\mathrm{vis}}\mathcal{L}_{\mathrm{vis}}+\lambda_{\mathrm{rt}}\mathcal{L}_{\mathrm{rt}}.
\end{equation}
Here, $\mathcal{L}_{\mathrm{mask}}$ combines binary cross-entropy and Dice losses for object segmentation.
$\mathcal{L}_{\mathrm{geo}}$ supervises object geometry, including boxes, centers, areas, mask--geometry consistency, and auxiliary slot attributes.
$\mathcal{L}_{\mathrm{vis}}$ supervises visibility labels derived from offline annotations; visibility is distinct from slot validity.
$\mathcal{L}_{\mathrm{rt}}$ is a bidirectional, multi-positive region--text contrastive loss between projected regional visual features and frozen CLIP text features.

\paragraph{VLA training.}
During VLA training, we keep the pretrained object encoder $E_\omega$ frozen and optimize the remaining trainable components with action imitation and auxiliary losses:
\begin{equation}
\label{eq:training_obj}
\mathcal{L}=w_{\mathrm{act}}\mathcal{L}_{\mathrm{act}}+\mathcal{L}_{\mathrm{aux}}, \qquad \mathcal{L}_{\mathrm{aux}}=\mathcal{L}_{\mathrm{cur}}+\mathcal{L}_{\mathrm{fut}}+\mathcal{L}_{\mathrm{prop}}+\mathcal{L}_{\mathrm{gate}}.
\end{equation}
Here, $\mathcal{L}_{\mathrm{act}}$ is the flow-matching action loss, which also optimizes the prospective object tokens through the action objective.
$\mathcal{L}_{\mathrm{cur}}$ supervises current object geometry, visibility, and mask logits, while $\mathcal{L}_{\mathrm{fut}}$ supervises future object states predicted by the forward dynamics module.
$\mathcal{L}_{\mathrm{prop}}$ trains the auxiliary coarse action proposal against the demonstrated action chunk, while $\mathcal{L}_{\mathrm{gate}}$ regularizes the learned slot-importance gates used for object-slot prioritization.
The routing projections and gates are optimized through the existing objectives without an additional routing loss.
$\mathcal{L}_{\mathrm{fut}}$ is computed only for valid non-robot object slots, excluding robot, invalid, and padded slots.

\begin{figure}
    \centering
    \includegraphics[width=0.95\linewidth]{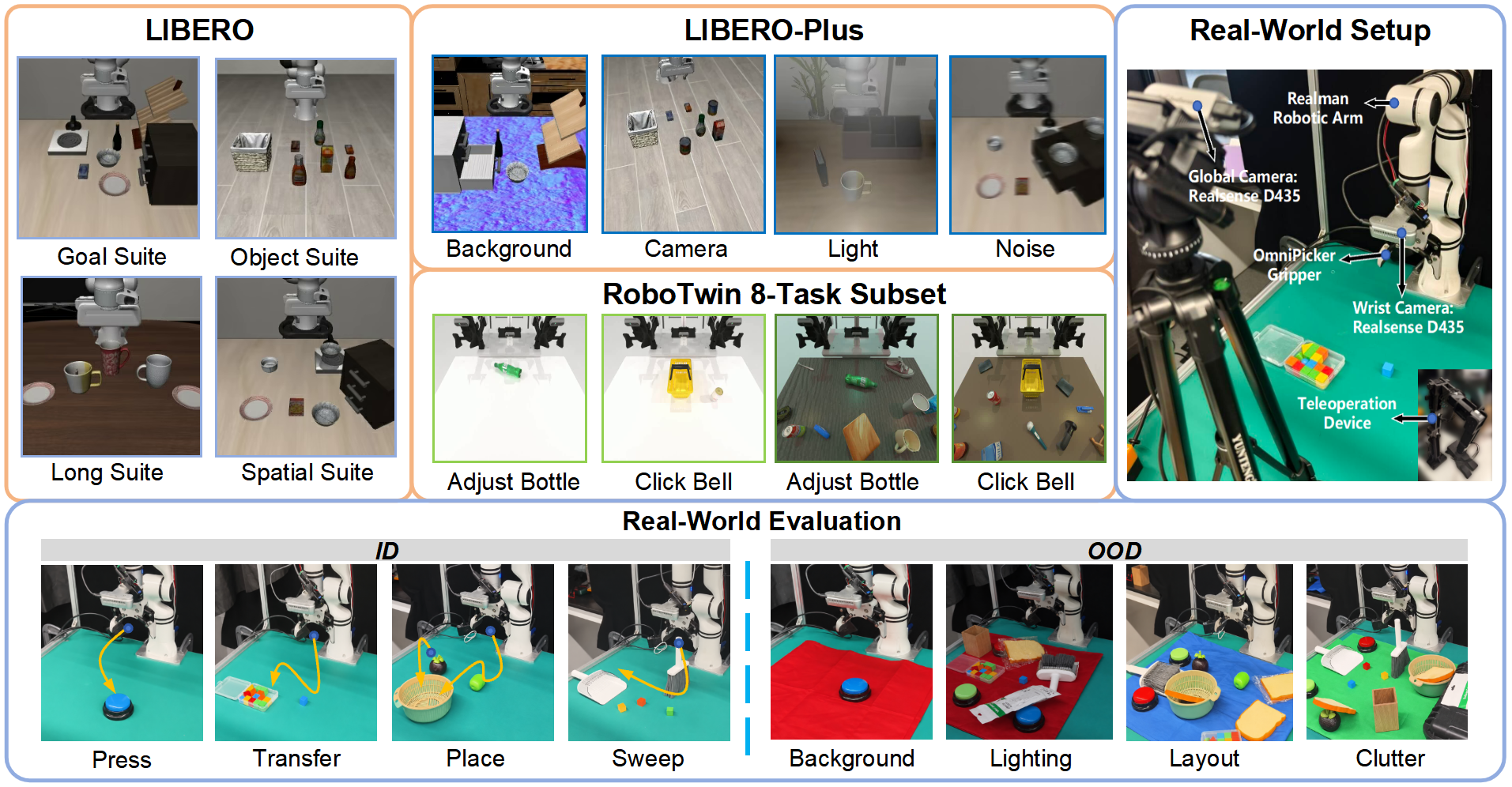}
    \caption{\textbf{Experimental overview.}
    We evaluate \ours{} on LIBERO, LIBERO-Plus, a fixed $8$-task subset of RoboTwin~2.0, and four real-world manipulation tasks, covering standard and perturbed evaluation settings.}
    \label{fig:experimental_setup}
\end{figure}

\section{Experiments}
\label{sec:experiments}

\paragraph{Benchmarks.}
We evaluate \ours{} on \libero{}, \libero-Plus, and RoboTwin~2.0, covering both single-arm and bimanual manipulation under diverse distribution shifts. For \libero{}, we follow the standard protocol over the Spatial, Object, Goal, and Long suites~\citep{liu2023libero}. We further evaluate zero-shot robustness on \libero-Plus under seven controlled shifts in camera viewpoint, robot initialization, object layout, language, lighting, background, and sensor noise~\citep{fei2025libero}.
For RoboTwin~2.0, we evaluate on a fixed 8-task subset: \textit{adjust bottle}, \textit{click bell}, \textit{grab roller}, \textit{move can pot}, \textit{open microwave}, \textit{place can basket}, \textit{scan object}, and \textit{turn switch}. These tasks cover diverse manipulation behaviors and object-state transitions. We report results under both Easy and Hard settings~\citep{chen2025robotwin}.

\paragraph{Implementation, training, and evaluation protocol.}
We use a Qwen3-VL-4B~\citep{bai2025qwen3} backbone and a layerwise flow-matching action expert, following the StarVLA training recipe~\citep{community2026starvla}.
Training proceeds in three stages: (1) we pretrain the spatio-temporal object encoder using offline object-level supervision; (2) with the encoder frozen, we jointly optimize the VLM, action expert, and policy-side object modules using Eq.~\ref{eq:training_obj}; and (3) we freeze the VLM and object-related modules and refine only the action expert with $\mathcal{L}_{\mathrm{act}}$.
Training is conducted on 8 NVIDIA H100 GPUs, while policy inference during evaluation is performed on NVIDIA RTX 4090 GPUs.
Each \libero{} task is evaluated over 50 episodes, with the same checkpoint evaluated on \libero-Plus without additional fine-tuning.
For RoboTwin~2.0, the policy is trained on 50 clean demonstrations per task and evaluated over 100 rollouts per task under both Easy and Hard settings.

\subsection{Main Results}
\paragraph{LIBERO and LIBERO-Plus.}
Standard \libero{} measures nominal manipulation performance, while \libero-Plus evaluates the same \libero{}-trained policy under controlled distribution shifts without additional fine-tuning.
As shown in Table~\ref{tab:libero_plus_main}, \ours{} achieves the highest average success rate on \libero{} (98.6\%) and ranks second overall on \libero-Plus (84.4\%), with the best results under language (90.6\%), lighting (98.4\%), and background (98.5\%) perturbations and second-best performance under sensor noise (88.6\%).
These results indicate strong robustness to semantic and appearance variations, while camera viewpoint and robot initialization remain the main challenges.

\begin{table*}[t]
\centering
\caption{\textbf{Performance on standard \libero{} and zero-shot robustness on \libero-Plus.}
All \libero-Plus results are evaluated without fine-tuning on perturbed environments.
Best results are shown in bold, and second-best results are underlined.}
\label{tab:libero_plus_main}
\scriptsize
\setlength{\tabcolsep}{2.0pt}
\begin{tabular}{lccccc@{\hspace{4pt}}|cccccccc}
\toprule
Method & \multicolumn{5}{c|}{\libero{}} & \multicolumn{8}{c}{\libero-Plus} \\
\cmidrule(lr){2-6}\cmidrule(lr){7-14}
 & Spatial & Object & Goal & Long & Avg & Cam. & Robot & Layout & Lang. & Light & BG & Noise & Overall \\
\midrule
\multicolumn{2}{l}{\textit{General VLA models}} & \multicolumn{12}{c}{} \\
OpenVLA~\citep{kim2024openvla}
& 84.7 & 88.4 & 79.2 & 53.7 & 76.5
& 0.8 & 3.5 & 28.5 & 23.0 & 8.1 & 34.8 & 15.2 & 15.6 \\
OpenVLA-OFT~\citep{kim2025fine}
& 97.6 & 98.4 & 97.9 & 94.5 & 97.1
& 56.4 & 31.9 & 74.2 & 79.5 & 88.7 & 93.3 & 75.8 & 69.6 \\
UniVLA~\citep{bu2025univla}
& 96.5 & 96.8 & 95.6 & 92.0 & 95.2
& 1.8 & 46.2 & 31.9 & 69.6 & 69.0 & 81.0 & 21.2 & 42.9 \\
$\pi_{0}$~\citep{black2024pi0}
& 96.8 & 98.8 & 95.8 & 85.2 & 94.2
& 13.8 & 6.0 & 68.9 & 58.8 & 85.0 & 81.4 & 79.0 & 53.6 \\
$\pi_{0.5}$~\citep{pi05}
& 98.8 & 98.2 & \underline{98.0} & 92.4 & 96.9
& \underline{75.4} & 77.5 & \textbf{85.7} & \underline{85.6} & \underline{96.9} & 94.6 & \textbf{89.7} & \textbf{85.7} \\
AVA-VLA~\citep{xiao2026ava}
& 97.4 & \underline{99.4} & 97.4 & \textbf{97.6} & 98.0
& 55.5 & 25.9 & 74.1 & \underline{85.6} & 95.5 & 88.9 & 78.0 & 70.1 \\
X-VLA~\citep{zheng2026x}
& 98.2 & 98.6 & 97.8 & \textbf{97.6} & \underline{98.1}
& 23.4 & \textbf{89.7} & 71.8 & 75.7 & 88.2 & \underline{96.0} & 62.7 & 71.4 \\
\midrule
\multicolumn{2}{l}{\textit{Object-centric and predictive models}} & \multicolumn{12}{c}{} \\
WorldVLA~\citep{cen2025worldvla}
& 87.6 & 96.2 & 83.4 & 60.0 & 81.8
& 0.1 & 27.9 & 38.0 & 41.6 & 43.7 & 17.1 & 10.9 & 25.0 \\
VLA-JEPA~\citep{sun2026vla}
& 96.2 & \textbf{99.6} & 97.2 & 95.8 & 97.2
& 63.3 & 67.1 & \underline{85.1} & 85.4 & 95.6 & 93.6 & 66.3 & 79.5 \\
OA-WAM~\citep{liu2026oa}
& \underline{98.9} & 99.0 & 97.4 & 95.9 & 97.8
& \textbf{80.5} & \underline{89.6} & 82.8 & 85.3 & 96.5 & 95.9 & 75.6 & 83.9 \\
\midrule
\textbf{\ours{} (Ours)}
& \textbf{99.2} & \textbf{99.6} & \textbf{98.2} & \underline{97.4} & \textbf{98.6}
& 66.2 & 75.8 & 81.2 & \textbf{90.6} & \textbf{98.4} & \textbf{98.5} & \underline{88.6} & \underline{84.4} \\
\bottomrule
\end{tabular}
\end{table*}

\begin{figure}[t]
    \centering
    \includegraphics[width=1\linewidth]{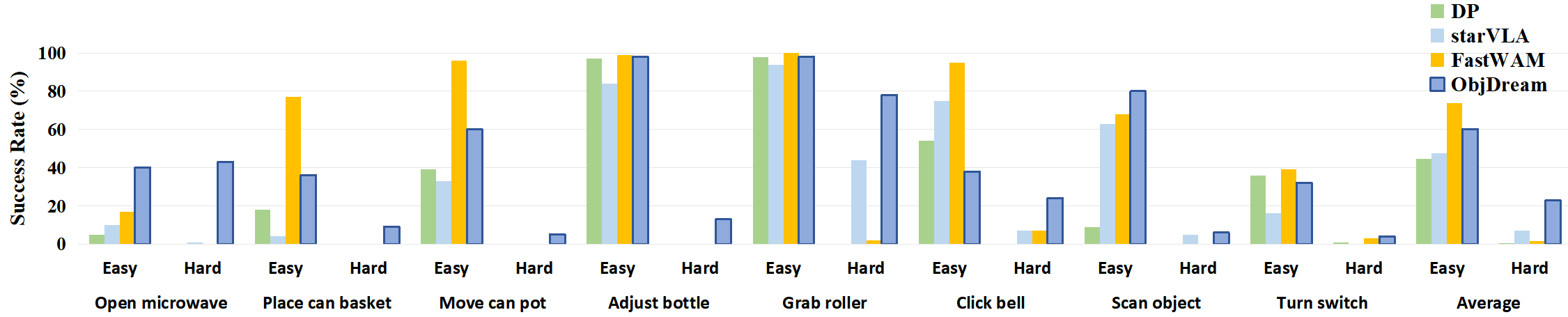}
    \caption{Success rates on the fixed $8$-task RoboTwin~2.0 subset under Easy and Hard settings.}
    \label{fig:robotwin_main}
\end{figure}

\paragraph{\textbf{RoboTwin~2.0.}}
As shown in Fig.~\ref{fig:robotwin_main}, we evaluate \ours{} on the fixed eight-task RoboTwin~2.0 subset under Easy and Hard settings.
\ours{} achieves average success rates of 60.25\% on Easy and 22.75\% on Hard, demonstrating competitive robustness under randomized scene variations without large-scale pretraining on robot data.
Compared with StarVLA, \ours{} improves the average success rate by 12.87 percentage points on Easy and 15.62 percentage points on Hard, with a larger gain under the more challenging Hard setting.
The improvement is particularly evident on \textit{grab roller}, \textit{open microwave}, and \textit{click bell} under Hard conditions.
These results indicate improved robustness to substantial distribution shifts in bimanual manipulation.

\subsection{Ablation and Analysis}

\begin{table*}[t]
    \centering
    \caption{\textbf{Ablation study of object-centric components.}
    Success rates (\%) on \libero-Long and \libero-Plus.
    Overall is the trial-weighted \libero-Plus average.
    Best results are shown in bold, and second-best results are underlined.}
    \label{tab:ablation_components}
    \footnotesize
    \setlength{\tabcolsep}{2.0pt}
    \begin{tabular}{lc@{\hspace{4pt}}|cccccccc}
        \toprule
        Variant
        & \multicolumn{1}{c|}{\libero{}}
        & \multicolumn{8}{c}{\libero-Plus} \\
        \cmidrule(lr){2-2}
        \cmidrule(lr){3-10}
        & Long
        & Cam. & Robot & Layout & Lang.
        & Light & BG & Noise & Overall \\
        \midrule
        Backbone baseline
        & 95.4
        & 66.4 & 74.7 & 80.0 & 87.1
        & 91.7 & 92.4 & 82.8 & 81.2 \\
        \midrule
        w/o ObjRel \& ObjDyn
        & 95.2
        & 65.6 & 73.7 & 78.9 & 87.4
        & 95.6 & 96.9 & 85.0 & 82.1 \\
        w/o ObjDyn
        & 96.2
        & \underline{66.6} & 75.0 & 80.6 & 88.2
        & 96.2 & 96.4 & 87.3 & 83.2 \\
        w/o ObjRel
        & \underline{96.6}
        & \textbf{67.4} & \textbf{76.1} & 80.9 & 88.2
        & \underline{97.6} & \underline{97.8}
        & \textbf{88.9} & \underline{84.1} \\
        w/o Fact. Injection
        & 96.2
        & 65.9 & 75.3 & \underline{81.0} & \underline{89.0}
        & 94.4 & 93.5 & 85.5 & 82.5 \\
        w/o Id. Routing
        & 95.8
        & 65.3 & 74.9 & 80.5 & 87.2
        & 95.3 & 96.4 & 86.6 & 82.6 \\
        \midrule
        \textbf{\ours{} (Full)}
        & \textbf{97.4}
        & 66.2 & \underline{75.8} & \textbf{81.2} & \textbf{90.6}
        & \textbf{98.4} & \textbf{98.5}
        & \underline{88.6} & \textbf{84.4} \\
        \bottomrule
    \end{tabular}
\end{table*}

We conduct ablations on \libero-Long and \libero-Plus to assess the object interface, object relation modeling (ObjRel), object dynamics (ObjDyn), factorized object injection (Fact. Injection), and identity-guided routing (Id. Routing).
Table~\ref{tab:ablation_components} includes a backbone baseline without the object interface, while ``w/o ObjRel \& ObjDyn'' retains the object tokens, factorized injection, identity-guided routing, and current state supervision.

The full model outperforms the backbone baseline by 2.0 and 3.2 percentage points on \libero-Long and \libero-Plus, respectively.
The object interface without ObjRel or ObjDyn provides a modest improvement over the backbone on \libero-Plus and comparable performance on \libero-Long.
Among single-component removals, disabling factorized injection or identity-guided routing produces the largest decreases in \libero-Plus performance, from 84.4\% to 82.5\% and 82.6\%, respectively.
Removing ObjDyn reduces the overall success rate to 83.2\%, while removing ObjRel results in a smaller decrease to 84.1\%.
Overall, factorized injection, identity-guided routing, and future object state supervision provide complementary gains.

\subsection{Real-World Experiments}

\label{sec:real_world}

\begin{table*}[t]
    \centering
    \caption{\textbf{Real-robot manipulation results.}
    Success rates (\%) on four real-world tasks under ID and OOD evaluation.
    Best and second-best results within each setting are bold and underlined.}
    \label{tab:real_world}
    \footnotesize
    \setlength{\tabcolsep}{3pt}
    \renewcommand{\arraystretch}{1.15}
    \begin{tabular*}{\textwidth}{@{\extracolsep{\fill}}ll|cccc|c@{}}
        \toprule
        Eval. & Method
        & \shortstack{Button\\Pressing}
        & \shortstack{Block\\Pick-and-Place}
        & \shortstack{Two-Fruit\\Placement}
        & \shortstack{Tabletop\\Sweeping}
        & \shortstack{Avg.\\Success} \\
        \midrule
        \multirow{3}{*}{ID}
        & $\pi_{0.5}$~\citep{pi05}
        & \textbf{90} & \textbf{70} & 40
        & \underline{65} & \underline{66} \\
        & ControlVLA~\citep{li2025controlvla}
        & 65 & 40 & \textbf{70}
        & \underline{65} & 60 \\
        & \textbf{\ours{} (Ours)}
        & \underline{80} & \underline{65} & \underline{60}
        & \textbf{90} & \textbf{74} \\
        \midrule
        \multirow{3}{*}{OOD}
        & $\pi_{0.5}$~\citep{pi05}
        & \textbf{65} & \textbf{60} & \underline{40}
        & \underline{45} & \underline{53} \\
        & ControlVLA~\citep{li2025controlvla}
        & \underline{40} & 10 & 35
        & 40 & 31 \\
        & \textbf{\ours{} (Ours)}
        & \textbf{65} & \underline{50} & \textbf{50}
        & \textbf{55} & \textbf{55} \\
        \bottomrule
    \end{tabular*}
\end{table*}

\paragraph{Real-world experiments.}
We evaluate \ours{} on four real-world tasks:
\textit{button pressing}, \textit{block pick-and-place},
\textit{two-fruit placement}, and \textit{tabletop sweeping},
covering contact interaction, object manipulation, and long-horizon control.
We compare against $\pi_{0.5}$~\citep{pi05} and ControlVLA~\citep{li2025controlvla}.
All methods use the same task demonstrations for adaptation and are evaluated under in-distribution (ID) and out-of-distribution (OOD) settings without additional fine-tuning.

\textbf{Results.}
As shown in Table~\ref{tab:real_world}, \ours{} achieves average success rates of 74\% and 55\% under ID and OOD conditions, respectively, exceeding the strong $\pi_{0.5}$ baseline by 8 percentage points under ID while achieving comparable aggregate performance under OOD.
Compared with ControlVLA, \ours{} improves average success rates by 14 and 24 percentage points under ID and OOD conditions, respectively.
Notably, under OOD conditions, \ours{} outperforms the best baseline by 10 percentage points on both \textit{two-fruit placement} and \textit{tabletop sweeping}, two tasks that require precise manipulation.
These results demonstrate competitive generalization in real-world manipulation without large-scale pretraining on robot data.

\section{Conclusion}
\label{sec:conclusion}

We present \ours{}, an object-centric VLA framework combining structured object representations with action-conditioned future object state supervision.
\ours{} represents objects with persistent identity cues and time-varying state features, and leverages identity-guided routing to direct VLM attention toward task-relevant object tokens.
It further supervises object state evolution during training without requiring the decoding of future object states at inference.
Experiments on \libero{}, \libero-Plus, a fixed $8$-task RoboTwin~2.0 subset, and four real-world tasks demonstrate strong manipulation performance and robustness to scene variations.

\clearpage

\FloatBarrier
\setlength{\bibsep}{4pt plus 0.3pt minus 0.3pt}
\bibliographystyle{plainnat}
\bibliography{references}

\clearpage
\appendix
\section{Offline Annotation Pipeline}
\label{app:offline_annotation}

\begin{figure}[H]
    \centering
    \includegraphics[width=\linewidth]{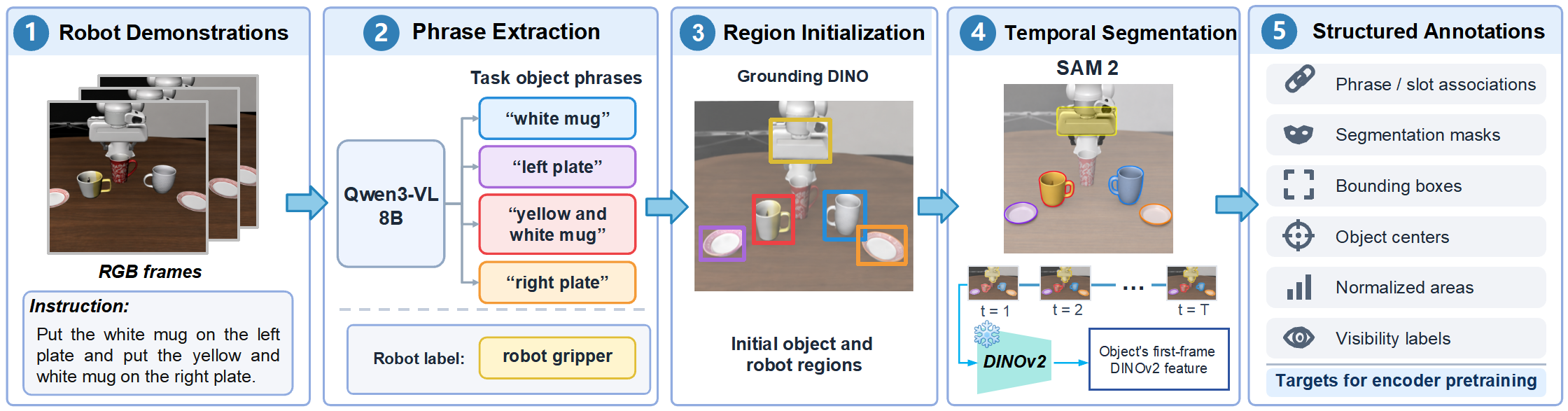}
    \caption{\textbf{Automatic offline annotation pipeline for object encoder pretraining.}
    Qwen3-VL-8B extracts task-relevant object phrases from demonstration instructions and initial observations, while a separate \emph{robot gripper} label is used for robot annotation.
    Grounding DINO initializes the corresponding object regions, SAM2 propagates segmentation masks across frames, and DINOv2 extracts object-specific features from the initial grounded regions.
    Masks, geometry, visibility, and phrase--region correspondences are automatically constructed from this pipeline without manual frame-wise object annotation.
    The resulting annotations supervise encoder pretraining, while the DINOv2 features provide visual identity references.}
    \label{fig:offline_annotation_pipeline}
\end{figure}

Importantly, the object-level supervision is automatically generated from the demonstration trajectories and does not require manual frame-wise annotation of masks, boxes, geometry, or visibility.
For each trajectory with instruction $x$ and visual observations, we construct object-level supervision through phrase extraction, grounded localization, mask propagation, and automatic object-state annotation, yielding temporally aligned masks, geometry, visibility, and phrase--region correspondences.
The offline annotation pipeline is illustrated in Fig.~\ref{fig:offline_annotation_pipeline}.

\paragraph{Instruction-conditioned object phrases.}
During automatic offline annotation, Qwen3-VL-8B~\citep{bai2025qwen3} takes the language instruction $x$ together with the initial visual observation $I_0$ and extracts the task-relevant phrase set $P_0$.
The extracted phrases are fixed for the entire trajectory and used as consistent semantic references across frames.
When multiple instances share the same category, coarse spatial cues, such as relative left/right positions, are included when needed to distinguish them.
During offline annotation, the label \emph{robot gripper} is used to localize robot regions and assign their robot role.

\paragraph{Phrase-grounded localization and temporal segmentation.}
Given $P_0$, Grounding DINO~\citep{groundingdino} provides initial phrase-conditioned localization, and SAM2~\citep{sam2} propagates the corresponding masks across the demonstration trajectory.
The phrase associated with each object remains fixed while its spatial support changes with motion, occlusion, or visibility.
Boxes, centers, areas, and visibility labels are automatically derived from the propagated masks and normalized with respect to the image dimensions where applicable.
DINOv2~\citep{dinov2} features are pooled over the initial grounded object region to construct the visual identity reference during encoder pretraining.
Thus, the masks, geometry, and visibility labels serve as automatically generated offline supervision and are not provided as policy inputs during rollout.
Grounding DINO localization and SAM2 video mask propagation are used only for offline annotation and are not executed during rollout.

\clearpage

\section{Future Object State Prediction Analysis}
\label{app:future_prediction_analysis}
\setlength{\intextsep}{4pt}
\setlength{\textfloatsep}{12pt plus 2pt minus 2pt}

We analyze whether the auxiliary future object state decoder learns nontrivial object state evolution rather than simply preserving the current state.
We evaluate three representative LIBERO manipulation tasks covering object placement and articulated-object interaction.
Only valid non-robot object slots are evaluated; robot and padding slots are excluded.
For each prediction origin, the decoder predicts object centers, bounding boxes, and masks at every consecutive future step $h\in\{1,\ldots,H_f\}$, with $H_f=8$.

We compare the learned decoder against a \emph{current state persistence} baseline that repeats the current object estimate at each future step.
We report an \textit{Unclamped} subset containing only predictions whose future targets are directly available within the trajectory, and a \textit{Displacement} subset containing object--target pairs whose centroid changes by more than $1\%$ of the image diagonal.
Center error is normalized by the image diagonal, while box and mask IoU measure predicted object extent.

\begin{table}[H]
    \centering
    \caption{\textbf{Future object state prediction accuracy on LIBERO.}
    Current state persistence repeats the current object state at each future target.
    Bold indicates the better value within each comparison.}
    \label{tab:app_future_prediction_results}
    \small
    \setlength{\tabcolsep}{4pt}
    \renewcommand{\arraystretch}{1.05}
    \begin{tabular}{@{}lllll@{}}
        \toprule
        \textbf{Subset}
        & \textbf{Method}
        & \begin{tabular}[c]{@{}l@{}}
            \textbf{Center}\\
            \textbf{error (\%)} $\downarrow$
          \end{tabular}
        & \textbf{Box IoU} $\uparrow$
        & \textbf{Mask IoU} $\uparrow$ \\
        \midrule

        Unclamped
        & Current state persistence
        & 6.37 & 0.634 & 0.564 \\
        & Future prediction
        & \textbf{1.60}
        & \textbf{0.690}
        & \textbf{0.616} \\

        \addlinespace[3pt]
        Displacement
        & Current state persistence
        & 13.91 & 0.287 & 0.245 \\
        & Future prediction
        & \textbf{2.03}
        & \textbf{0.575}
        & \textbf{0.542} \\
        \bottomrule
    \end{tabular}
\end{table}

The learned decoder consistently improves over current state persistence.
On the Unclamped subset, the learned decoder has lower center error and higher box and mask IoU than current state persistence.
The gains are more pronounced under object displacement, where center error decreases from $13.91\%$ to $2.03\%$ and box and mask IoU improve from $0.287$ and $0.245$ to $0.575$ and $0.542$.
These results indicate that the auxiliary decoder captures nontrivial future object state changes rather than merely propagating the current state.

\begin{figure}[!t]
    \centering
    \includegraphics[width=0.9\linewidth]{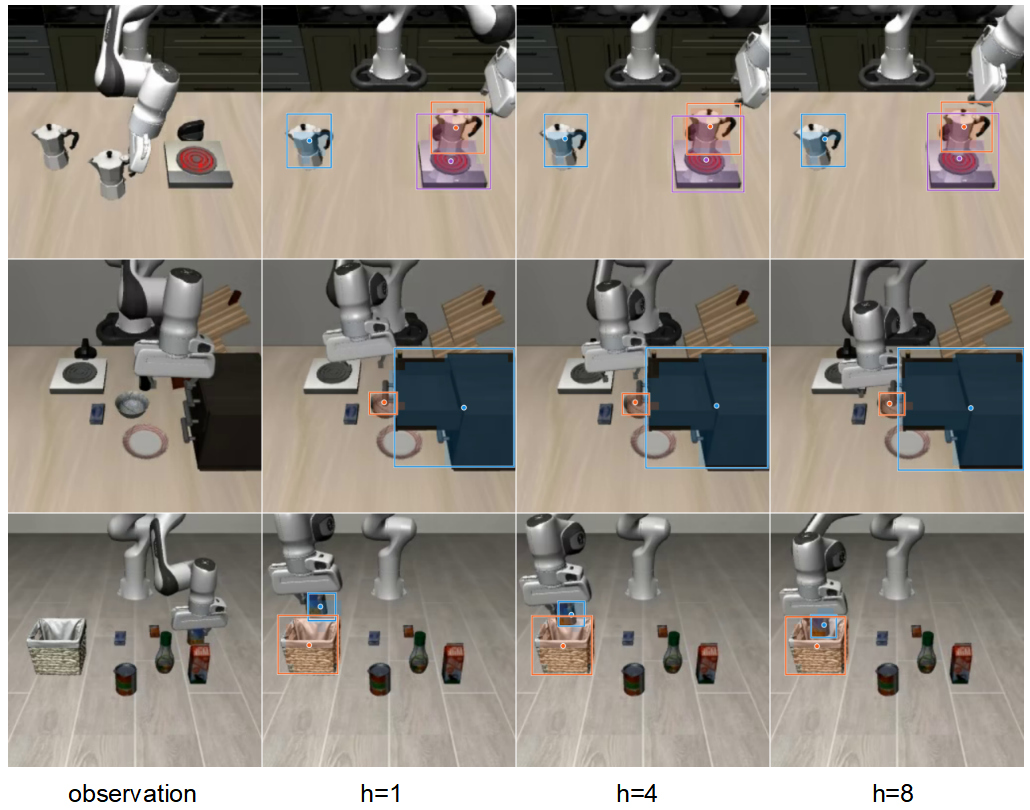}
    \vspace{0pt}
    \setlength{\abovecaptionskip}{0pt}
    \caption{\textbf{Qualitative future object mask predictions on LIBERO.}
    Rows show three representative manipulation tasks.
    The decoder predicts all future steps $h=1,\ldots,H_f$ for non-robot objects; for clarity, we visualize only $h=1,4,8$.
    The predicted masks are overlaid on the corresponding future RGB frames, which are used only for visualization and are not provided to the decoder.}
    \vspace{-5pt}
    \label{fig:app_future_reconstruction_examples}
\end{figure}

\begin{figure}[!t]
    \centering
    \includegraphics[width=0.9\linewidth]{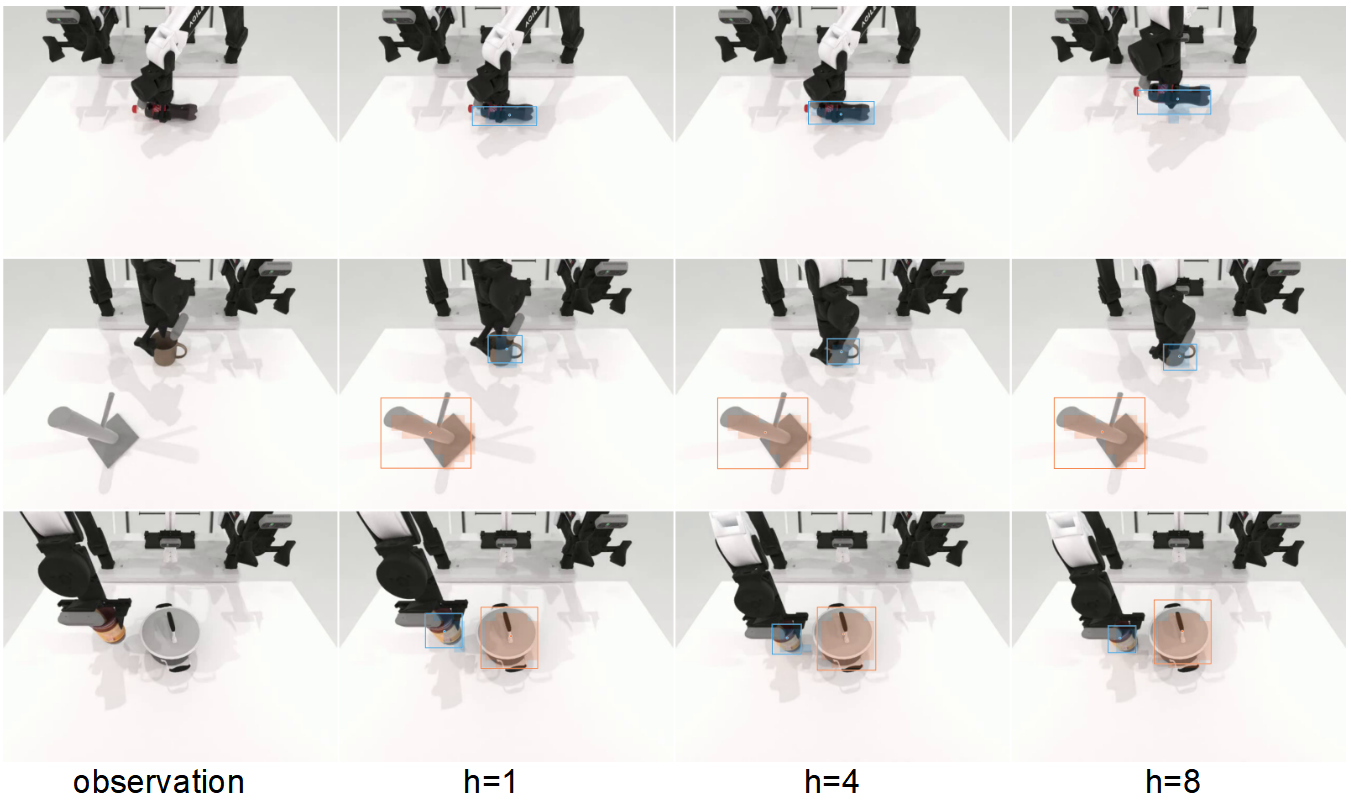}
    \setlength{\abovecaptionskip}{0pt}
    \caption{\textbf{Qualitative future object state predictions on RoboTwin~2.0.}
    Rows show representative tasks from the fixed eight-task subset.
    The decoder predicts all future steps for non-robot objects; for clarity, we visualize masks and bounding boxes only at $h=1,4,8$.
    The predictions are overlaid on the corresponding future RGB frames, which are used only for visualization and are not provided to the decoder.}
    \vspace{-5pt}
    \label{fig:app_future_robotwin}
\end{figure}

\begin{figure}[!t]
    \centering
    \includegraphics[width=0.9\linewidth]{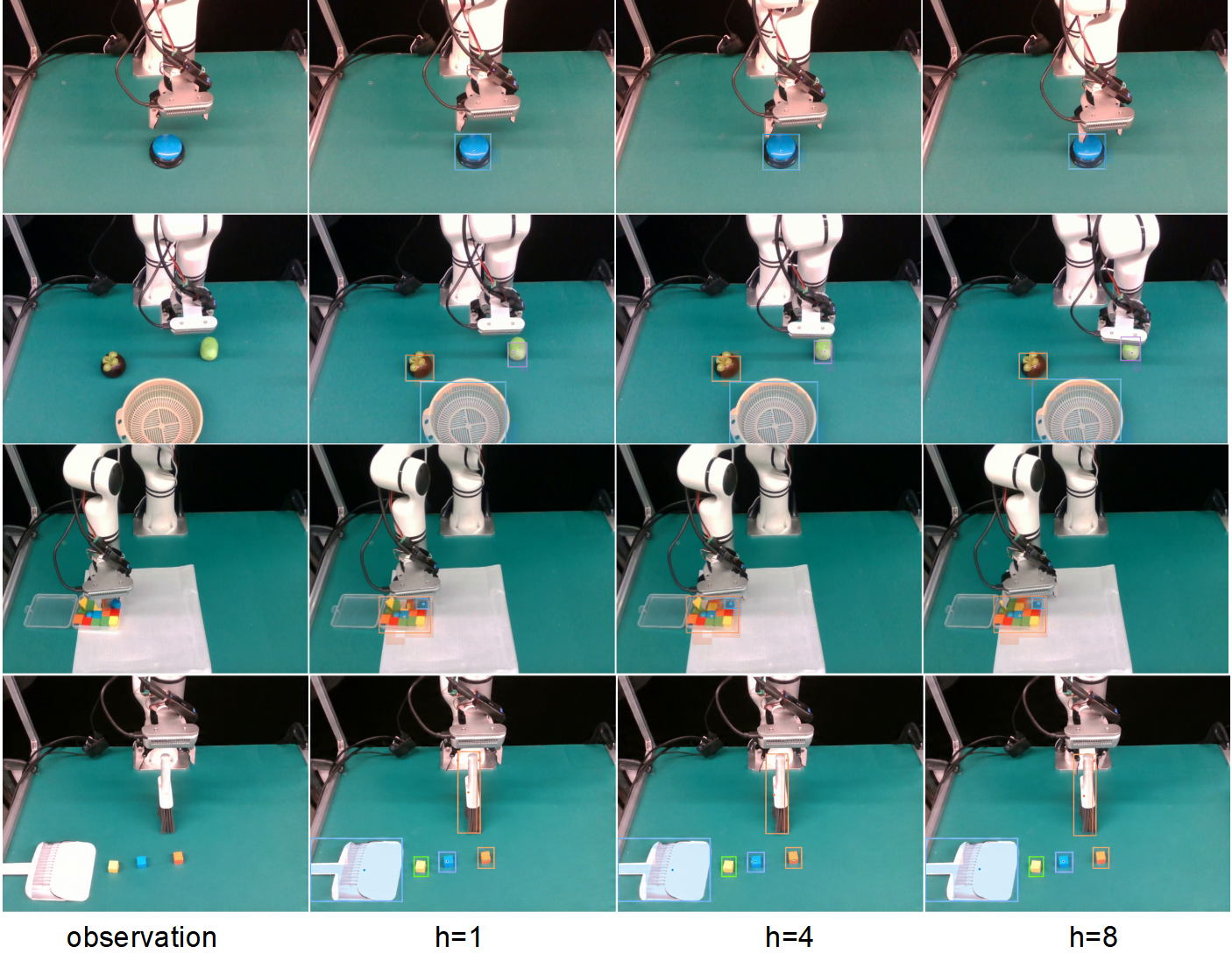}
    \setlength{\abovecaptionskip}{0pt}
    \caption{\textbf{Qualitative future object state predictions on real-world tasks.}
    Rows show button pressing, two-fruit placement, block pick-and-place, and tabletop sweeping.
    The decoder predicts all future steps for non-robot objects; for clarity, we visualize masks and bounding boxes only at $h=1,4,8$.
    The predictions are overlaid on the corresponding future RGB frames, which are used only for visualization and are not provided to the decoder.}
    \vspace{-5pt}
    \label{fig:app_future_realworld}
\end{figure}

\paragraph{Qualitative future object state predictions.}
Figure~\ref{fig:app_future_reconstruction_examples} shows representative LIBERO future object state predictions.
Although the decoder predicts all horizons $h=1,\ldots,H_f$, we display only $h=1,4,8$ for clarity.
Figures~\ref{fig:app_future_realworld} and~\ref{fig:app_future_robotwin} show additional predictions on real-world tasks and RoboTwin~2.0, respectively.

\begin{table}[H]
\centering
\caption{
\textbf{Sensitivity of future object state prediction to action conditioning on LIBERO.}
\textit{Native Predicted Action} uses the original auxiliary action proposal;
\textit{No Action} removes the position-aware action embedding; and
\textit{Mismatched Action} replaces it with the position-aware embedding of a predicted action chunk from another sample within the same task.
All other model components and prediction targets are kept fixed.
}
\label{tab:action_condition_sensitivity}
\small
\begin{tabular}{llccc}
\toprule
Subset & Conditioning
& Center error (\%) $\downarrow$
& Box IoU $\uparrow$
& Mask IoU $\uparrow$ \\
\midrule
\multirow{3}{*}{Unclamped}
& No Action
& 1.90 & 0.659 & 0.609 \\
& Mismatched Action
& 2.02 & 0.660 & 0.587 \\
& Native Predicted Action
& \textbf{1.60} & \textbf{0.690} & \textbf{0.616} \\
\midrule
\multirow{3}{*}{Displacement}
& No Action
& 2.35 & 0.546 & 0.534 \\
& Mismatched Action
& 2.71 & 0.532 & 0.486 \\
& Native Predicted Action
& \textbf{2.03} & \textbf{0.575} & \textbf{0.542} \\
\bottomrule
\end{tabular}
\end{table}

\paragraph{Effect of action conditioning on future object state prediction.}
To examine whether the decoder uses action information, we intervene on the action conditioning path using the same trained model and evaluation tasks.
Model parameters, contextualized object tokens, and action conditioning gates are kept fixed.
\textit{Native Predicted Action} uses the original auxiliary action proposal;
\textit{No Action} sets the position-aware action embedding to zero; and
\textit{Mismatched Action} replaces it with the position-aware embedding of a predicted action chunk from another sample within the same task.
As shown in Table~\ref{tab:action_condition_sensitivity}, the native condition yields the lowest center error and the highest box and mask IoU on both subsets.
On the Unclamped subset, center error is $1.60\%$ with the native condition, compared with $1.90\%$ without action conditioning and $2.02\%$ with mismatched actions.
On the Displacement subset, center error increases from $2.03\%$ to $2.35\%$ and $2.71\%$, respectively.
Box IoU decreases from $0.575$ to $0.546$ and $0.532$, while mask IoU decreases from $0.542$ to $0.534$ and $0.486$.
These interventions indicate that the future object state decoder uses the action condition, while not establishing counterfactual dynamics under arbitrary action sequences.

\end{document}